\documentclass[pmlr,twocolumn,10pt]{jmlr}

\mlhtrack{proceedings}

\newif\iffinal
\finaltrue  

\iffinal
    \ifmlhneedspmlr
      \jmlrvolume{346}
      \jmlryear{2026}
    \fi
    \ifmlhfindings \jmlrproceedings{}{ML4H 2026 - Findings Track}\fi
    \ifmlhdemo     \jmlrproceedings{}{ML4H 2026 - Demo Track}\fi
    \jmlrworkshop{Machine Learning for Health (ML4H) 2026}
\else
    \jmlrproceedings{}{Submitted to ML4H 2026: \mlhtrackname}
    \jmlrworkshop{Machine Learning for Health (ML4H) 2026}
\fi

\usepackage{float}
\usepackage{booktabs}
\usepackage{siunitx}
\usepackage{multirow}
\usepackage{array}
\usepackage{listings}
\usepackage{tcolorbox}
\tcbuselibrary{breakable,listings}

\usepackage[switch]{lineno}

\newtcblisting{promptbox}[1]{
  breakable,
  listing only,
  colback=gray!5,
  colframe=gray!40!black,
  title={#1},
  fonttitle=\bfseries,
  boxrule=0.5pt,
  arc=1mm,
  left=1mm,
  right=1mm,
  top=1mm,
  bottom=1mm,
  listing options={
    basicstyle=\ttfamily\scriptsize,
    breaklines=true,
    columns=fullflexible,
    keepspaces=true,
    showstringspaces=false
  }
}

\makeatletter
\long\def\@makecaption#1#2{%
\vskip\abovecaptionskip
{\small #1: #2\par}%
\vskip\belowcaptionskip}
\makeatother

\newcommand{\omitehr}{OmitEHR}
\newcommand{\fsal}{f_{\mathrm{sal}}}
\newcommand{\fcov}{f_{\mathrm{cov}}}

\title[\omitehr]{\omitehr{}: Scalable Omission Detection in LLM-Generated Clinical Summaries}

\author{%
\Name{Bridget Lin}\textsuperscript{1,*} \Email{bridget7@stanford.edu}\\
\Name{Ryan Welch}\textsuperscript{2,*} \Email{rcwelch@stanford.edu}\\
\Name{Bobak Seddighzadeh}\textsuperscript{3} \Email{bobaks@stanford.edu}\\
\Name{Fran Riley}\textsuperscript{3} \Email{franr@stanford.edu}\\
\Name{Kaitlyn Stoehr}\textsuperscript{3} \Email{kstoehr@stanford.edu}\\
\Name{Margarita I. Pipinos}\textsuperscript{3} \Email{mpipinos@stanford.edu}\\
\Name{Dimitri Speron}\textsuperscript{3} \Email{dim2552@stanford.edu}\\
\Name{Emma V. Sage}\textsuperscript{3} \Email{emmsage@stanford.edu}\\
\Name{Sneha Jain}\textsuperscript{3} \Email{snehajain@stanford.edu}\\
\Name{Nigam H. Shah}\textsuperscript{1,2,3,\dag} \Email{nigam@stanford.edu}\\
\Name{Emily Alsentzer}\textsuperscript{1,2,3,\dag} \Email{ealsentzer@stanford.edu}\\
\addr
Departments of \textsuperscript{1}Biomedical Data Science, \textsuperscript{2}Computer Science, and \textsuperscript{3}Medicine, Stanford University, USA\\[3pt]
{\normalfont\footnotesize
\textsuperscript{*}Equal contribution (alphabetical order).\quad
\textsuperscript{\dag}Equal supervision.}
}

\begin{document}

\maketitle
\jmlrauthors{B. Lin, R. Welch, B. Seddighzadeh,
F. Riley, K. Stoehr, M. I. Pipinos,
D. Speron, E. V. Sage, S. Jain,
N. H. Shah \& E. Alsentzer}

\ifmlhdemo\else
\begin{abstract}
Health systems increasingly rely on large language models (LLMs) to draft summaries of patient medical records, yet evaluations now find omissions, the absence of clinically important information, to be the most common failure mode. Detecting omissions in an LLM-generated summary requires knowing which facts in the record are salient, a signal that existing methods hand-specify, ignore, or recompute at inference with costly LLM judges. We present \omitehr{}, a reference-free omission detector that learns clinical salience from historical clinician-written summaries predating widespread LLM use. \omitehr{} learns to predict which source-note sentences clinicians include in their summaries, then surfaces high-salience sentences not covered by an LLM-generated summary as candidate omissions. We apply \omitehr{} to brief hospital course (BHC) summarization using 5,000 admissions from a large academic medical center. On held-out admissions, \omitehr{} outperformed all evaluated baselines at predicting salient source sentences, including a prompted LLM judge. The full omission detection pipeline took 17.7 to 19.5 seconds and cost \$0.03 to \$0.13 per summary depending on the base model, 1.6 to 1.7 times faster and 5.6 to 6.7 times cheaper than scoring every source sentence with an LLM judge. In a blinded review of 25 admissions, two board-certified clinicians confirmed a mean of 5.6 clinically meaningful omissions per summary surfaced by \omitehr{}, compared with 3.3 for a full LLM pipeline, counting candidates confirmed by at least one clinician. The efficiency and validity of \omitehr{} make it a practical tool for health systems to flag omissions before a summary reaches the clinician.
\end{abstract}

\fi

\ifmlhneedsstatements
\paragraph*{Data and Code Availability}
This work uses de-identified EHR data from an academic medical center. The clinical notes may contain protected health information and cannot be shared publicly. Code for the labeling pipeline, salience model, and omission detector is available at \url{https://github.com/alsentzerlab/omitehr}.
\paragraph*{Institutional Review Board (IRB)}
This study was conducted using de-identified electronic health record data under a protocol approved by the Stanford University Institutional Review Board (protocol \#57916).
\fi

\section{Introduction}
\label{sec:intro}

\begin{figure*}[t]
  \centering
  \includegraphics[width=\textwidth]{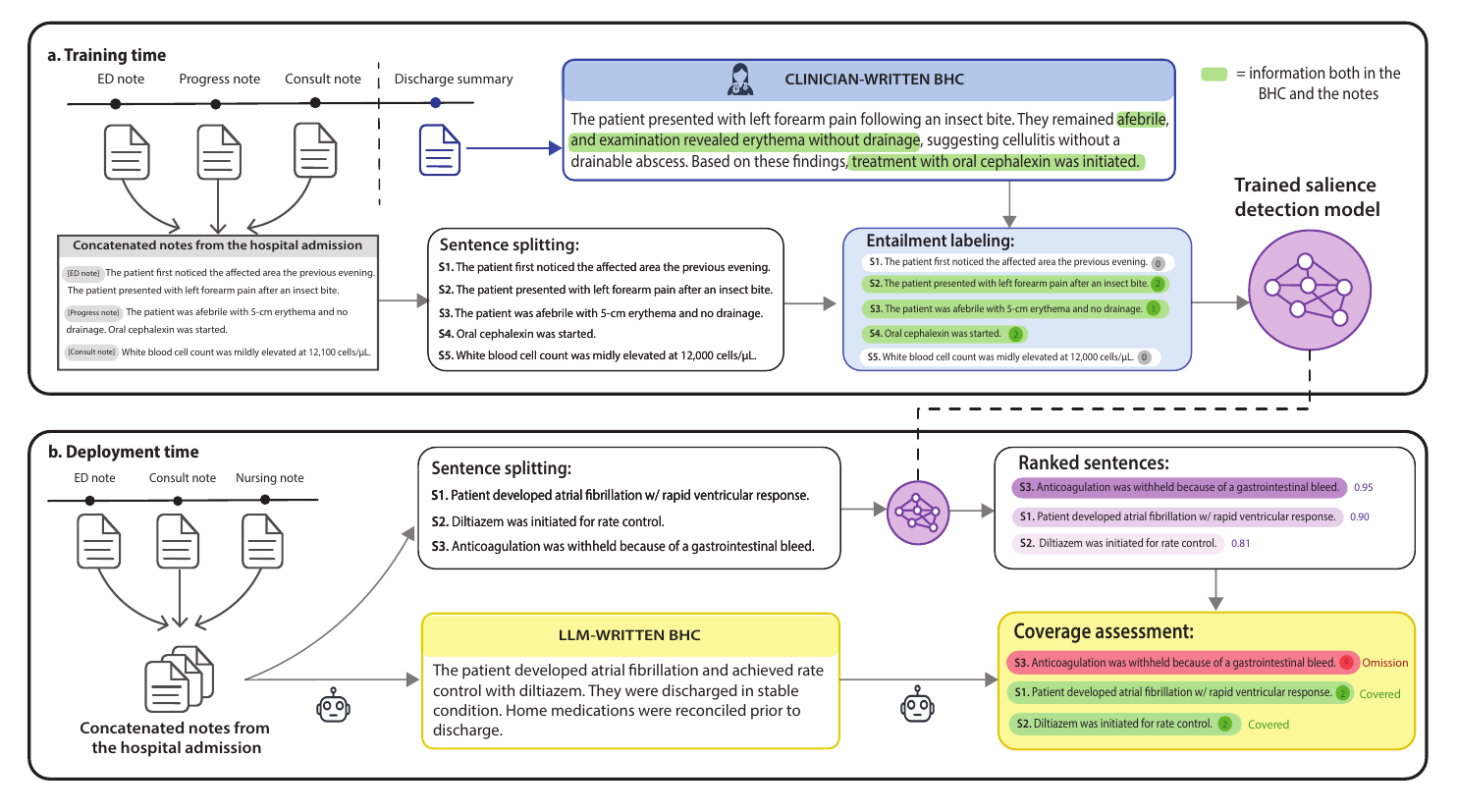}
  \caption{The \omitehr{} pipeline. Labels indicate content absent from (0), indirectly covered by (1), or directly stated in (2) the BHC. (a)~At training time, an LLM aligns each source-note sentence to the clinician-written BHC, and the resulting inclusion labels train a contextual salience model. (b)~At deployment, the model ranks the sentences of a new admission by salience, only the highest-ranked sentences are checked for coverage against the LLM-written BHC, and uncovered sentences are surfaced as candidate omissions. Examples are illustrative.}
  \label{fig:pipeline}
\end{figure*}

Chart review and documentation occupy a substantial share of clinicians' time, with studies finding that nearly half is consumed by desk work \citep{sinsky2016allocation, tierney2024ambient, arndt2017tethered}. In response, health systems are increasingly deploying tools powered by large language models (LLMs) that summarize the electronic health record (EHR) at the point of care, including Epic's integrated chart summarization \citep{kahl2026evaluation}, Stanford's ChatEHR \citep{shah2026chatehr}, and Penn Medicine's Chart Hero \citep{pennmedicine2026charthero}. Although LLM-generated summaries may improve clinician productivity, they exhibit two key failure modes: 1)~\emph{hallucinations}, statements in the summary not supported by the patient's record, and 2)~\emph{omissions}, clinically important information from the record that is missing from the summary. Existing research primarily targets hallucinations, where each claim in a summary can be evaluated for support against the source \citep[e.g.,][]{chung2025verifact}. However, clinician evaluations identify omissions as the more frequent failure mode. Across 147 AI-generated chart summaries, clinicians flagged 46 omissions and only 5 hallucinations \citep{kahl2026evaluation}, and in a separate evaluation of GPT-4 emergency department summaries, 47\% of summaries contained omissions and 42\% hallucinations \citep{williams2025evaluating}.

Omissions can disrupt handoffs, delay care, and increase the risk of adverse events \citep{smith2005missing, kripalani2007deficits}, and they are harder to detect than hallucinations because there is no fixed set of claims to verify. A detector must review the patient chart, decide which facts are salient, and then determine whether the summary covers them. Salience is central to omission detection, yet existing methods either ignore it or estimate it with LLM judges at inference time \citep{oukelmoun2025detecting, schumacher2025medomit, bedi2026care}. Judging salience with an LLM is expensive, as it requires reviewing the entire source record, and frequently diverges from physicians' judgments of relevance \citep{hao2025medpairmeasuringphysiciansai}. LLM services charge by token usage, so repeatedly processing full records makes omission detection more expensive as records grow longer. As a result, health systems now monitor token spending as a budget line and discontinue AI tools whose cost exceeds their value \citep{diaz2026tokens}.

Supervision to train a clinical salience model, however, already exists at scale. Clinicians have written summaries of hospitalizations for decades, and these summaries provide evidence of which information they considered worth including. Across thousands of admissions, these inclusion decisions provide a training signal for clinical salience that requires no manual annotation.

We introduce \omitehr{}, a reference-free omission detector that learns clinical salience from these historical inclusion decisions (Figure~\ref{fig:pipeline}). \omitehr{} separates omission detection into two steps, identifying salient information in the source record and determining whether the generated summary covers it. For the first step, a model trained on historical record--summary pairs predicts, for a new admission, which source sentences a clinician's summary would have covered, effectively predicting the content of the reference summary that is unavailable at deployment. For the second, an LLM judge checks only the highest-ranked sentences for coverage by the generated summary, and uncovered sentences are surfaced as candidate omissions. Since salience is learned offline, the only LLM computation at deployment is a fixed number of coverage checks, so the cost of detection does not grow with the length of the record.

We apply \omitehr{} to brief hospital course (BHC) summarization, which synthesizes documentation from across a hospitalization into the narrative hospital-course section of the discharge summary \citep{adams2021whats}. Using 5,000 admissions comprising 2.9 million source sentences, we train a bidirectional Transformer over source-sentence embeddings to predict each sentence’s salience in the context of the full admission. Training labels indicate whether each source sentence is represented in the clinician-written BHC. At inference, \omitehr{} screens an LLM-generated BHC for omissions in 17.7 to 19.5 seconds and at a cost of \$0.03 to \$0.13 on average, 1.6 to 1.7 times faster and 5.6 to 6.7 times cheaper than judging every source sentence with an LLM. In a blinded review of 25 admissions, \omitehr{} surfaced a mean of 5.64 omissions per summary confirmed by at least one clinician, compared with 3.32 for a full LLM pipeline.

This paper contributes (1) a reframing of reference-free omission detection as learned clinical salience followed by summary-coverage assessment, (2) a scalable pipeline for deriving salience labels from historical clinician-written BHCs and a contextual salience model that outperforms structural, lexical, and prompted LLM-judge baselines, and (3) a low-cost omission detection pipeline whose surfaced omissions clinicians confirmed in a blinded review. 

\section{Related Work}
\label{sec:background}

\paragraph{Reference-free omission detection.}
Omissions in LLM-generated clinical summaries have generally been identified through manual clinician review \citep{kahl2026evaluation, williams2025evaluating} or by comparison against a clinician-written reference, but manual review does not scale and the reference is unavailable at deployment, since it is exactly the document the LLM was asked to produce. Reference-free methods instead detect omissions from the source record and the generated summary alone. EmbedKDECheck \citep{oukelmoun2025detecting} embeds source and summary text and uses kernel density estimation to flag regions of the source the summary does not cover, which is inexpensive but does not model clinical importance. MED-OMIT \citep{schumacher2025medomit} extracts atomic facts from dialogue, flags those an LLM judge finds excluded from a generated SOAP note, and weights each by its effect on a simulated differential diagnosis as a proxy for importance. The most closely related formulation is CARE \citep{bedi2026care}, which uses LLM judges at inference time to assign a clinical importance and summary coverage score to each source sentence, and calibrates thresholds on both scores with conformal risk control. Scoring every source sentence for every summary is expensive, and LLM estimates of clinical relevance frequently diverge from those of physicians \citep{hao2025medpairmeasuringphysiciansai}. \omitehr{} adopts the same importance-and-coverage decomposition as CARE but learns salience from historical clinician-written BHCs, so no LLM importance judgments are made at inference and salience is grounded in observed clinician decisions rather than in a judge's reading of a prompt.

\paragraph{Learning salience from clinician behavior.}
Clinician behavior has been utilized to provide a relevance signal for related tasks. \citet{jiang2024machine} derive relevance labels from EHR audit logs, treating the notes a clinician opened before documenting as evidence of notes worth reading. In general-domain summarization, content selection is classically learned from sentence- or entity-level inclusion labels obtained from reference summaries \citep{nallapati2017summarunner, liu2019text, lin2025gumsage}. SPEER \citep{adams2024speer} applies this idea to BHC summarization, labeling medical entities from the source note as salient when they appear in the reference BHC and training a classifier whose predictions guide a fine-tuned LLM's sentence planning. \omitehr{} uses the same supervision but differs in unit and purpose. It predicts salience over whole sentences rather than isolated terms, and it audits summaries rather than generating them, so the salience model can check the output of any summarizer for omitted content.

\section{Problem Formulation}
\label{sec:formulation}
Our goal is to identify clinically important information from the source notes of a hospitalization that a corresponding LLM-generated summary fails to include. Formally, let $\mathcal{S} = (s_1, \ldots, s_n)$ denote the sentences from an admission's source record, and let $\hat{B}$ be an LLM-written summary of that record. \omitehr{} identifies an \emph{omission} separately from two judgments: (i) whether $s_i$ is \emph{salient}, meaning it belongs in the summary, and (ii) whether it is \emph{covered}, meaning its content appears in $\hat{B}$.

We define the salience of $s_i$ as the probability that a clinician would include its clinical content in the summary, $\fsal(s_i \mid \mathcal{S}) \in [0,1]$, conditioned on the entire record. We learn this function from an offline set of historical clinician-written summaries $\mathcal{B}$. For each source record and its summary $B$, each source sentence receives an inclusion label $\ell_i \in \{0, 1, 2\}$ indicating whether its content is absent from, indirectly reflected in, or directly stated in $B$. Treating labels 1 and 2 as positive, we train $\fsal$ to predict $\mathbb{P}\left( \ell_i > 0 \mid \mathcal{S} \right)$.

We define the coverage of a sentence $s_i$ as its inclusion label in an LLM-generated summary, $\fcov(s_i, \hat{B}) \in \{0, 1, 2\}$, using the same inclusion scale. Coverage is a property of a specific sentence--summary pair and must be computed at inference. \omitehr{} retains the $K$ most salient sentences from the source records, defined as $\operatorname{TopK}\!\left(\mathcal{S}; \fsal\right)$, and surfaces those not covered by $\hat{B}$. The tunable budget $K$ controls the tradeoff between omission recall, review burden, and inference cost. The resulting set of candidate omissions is defined as:

\vspace{-1em}

\begin{equation*}
\label{eq:omissions}
\mathcal{O}(\hat{B}) = \operatorname{TopK}\!\left(\mathcal{S}; \fsal\right) \cap \left\{ s \in \mathcal{S} : \fcov(s, \hat{B}) = 0 \right\}
\end{equation*}

\section{Methods}
\label{sec:methods}

\begin{figure*}[t]
  \centering
  \includegraphics[
    width=0.8\textwidth,
    angle=0,
    origin=c,
    keepaspectratio
  ]{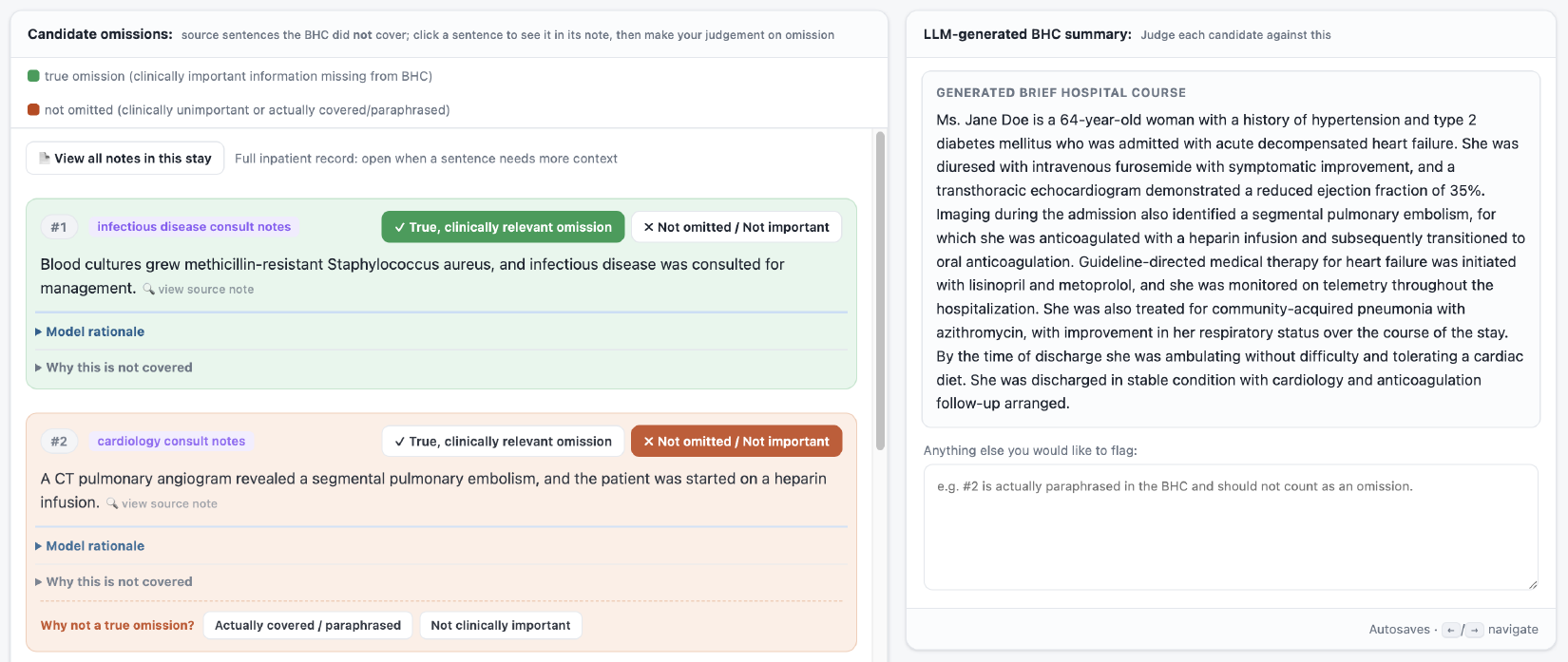}
  \caption{Example candidate omissions with completed fields: source sentence and note type, verdict buttons, LLM-generated rationale, and the coverage judge's reason. Patient content is synthetic.}
  \label{fig:ui_mini}
\end{figure*}

\subsection{Dataset Construction}
\label{methods:dataset}
We assemble a dataset of 5,000 internal medicine admissions at a large academic medical center. We restrict the cohort to stays longer than 24 hours, where longitudinal summarization is more useful. For each admission, we concatenate all notes from the hospitalization, excluding the discharge summary, to form the source. Notes are also the primary input to deployed summarization systems, though the exact input construction may vary \citep{shah2026chatehr, kahl2026evaluation, pennmedicine2026charthero}. The target summary is the brief hospital course (BHC) section of the discharge summary. Whereas other sections are often templated, the BHC is a clinician-written narrative of the admission’s key events and therefore provides naturally occurring supervision for which source information clinicians considered salient. Additional dataset characteristics are reported in Appendix~\ref{apd:dataset}.


\paragraph{Sentence segmentation.}
We segment the source notes into sentences using the medspaCy clinical sentence splitter \citep{medspacy}. We then apply rule-based postprocessing that removes templated content \citep{cahoon2026trace} and splits bulleted and numbered list items, which medspaCy often merges, into separate segments. We use sentences rather than LLM-extracted atomic facts because segmentation is deterministic, auditable, and preserves the note's wording, whereas decomposition requires an LLM call over every note and can drop or alter clinical details. 

\paragraph{Salience labels.}
\label{methods:labels}
Given each source sentence, its source note, and the clinician-written BHC, we prompt GPT-5-mini to assign one of three  labels describing how the sentence's content is reflected in the BHC (Table~\ref{tab:rubric}). Across the 2.9 million sentences, 63.5\% are labeled not related, 30.5\% indirectly related, and 5.9\% directly included (Appendix~\ref{apd:dataset}).

\begin{table}[t]
  \centering
  \small
  \setlength{\tabcolsep}{4pt}
  \renewcommand{\arraystretch}{1.1}
  \begin{tabular}{@{}>{\raggedright\arraybackslash}p{0.19\columnwidth}>{\raggedright\arraybackslash}p{0.40\columnwidth}>{\raggedright\arraybackslash}p{0.35\columnwidth}@{}}
    \toprule
    \multicolumn{3}{@{}>{\raggedright\arraybackslash}p{\dimexpr0.94\columnwidth+4\tabcolsep\relax}@{}}{\textbf{BHC excerpt.} ``Etiology of symptoms include bronchiectasis flare, PNA, or viral process, with the first most likely.''} \\
    \midrule
    Label & Definition & Source-note example \\
    \midrule
    Not related (0) & The BHC makes no claim this sentence supports. & ``All other systems reviewed and are negative.'' \\
    \addlinespace
    Indirectly related (1) & Supports a BHC claim but adds a value, timing, or method the BHC does not state. & ``Chronic bronchiectasis, presented with 3 days of worsening SOB.'' \\
    \addlinespace
    Directly included (2) & States the same fact as the BHC, in the same or different words. & ``Etiology of symptoms include bronchiectasis flare, PNA, or viral process.'' \\
    \bottomrule
  \end{tabular}
\caption{Inclusion-label rubric used for training and inference. \emph{Examples.} The review of systems supports no BHC claim (0). The bronchiectasis history relates to the BHC's etiology claim but adds history and symptom timing the BHC does not state (1). The etiology line states the same candidate causes as the BHC (2).}
\label{tab:rubric}
\end{table}

To validate the labeling scheme, two internal medicine clinicians independently annotated 300 sentences from 15 admissions using the same rubric, blinded to the LLM labels and to each other, with each sentence shown alongside its source note and the BHC. A third annotator adjudicated disagreements. The two clinicians agreed with each other at a linearly weighted Cohen's kappa of 0.647, and GPT-5-mini's labels showed moderate agreement with each clinician (0.454 and 0.414). Appendix~\ref{apd:entailment} details the labeling pipeline, annotation interface, and full agreement results.

\subsection{Salience Model}
\label{methods:model}

We train a class of models to predict the salience of each source sentence. We collapse the three inclusion labels of  Table~\ref{tab:rubric} into a binary target, treating indirectly related and directly included sentences as salient and not related sentences as non-salient.

Our primary model, \emph{\omitehr{}-contextual}, represents each sentence as its mean-pooled BioClinRoBERTa token embeddings \citep{lewis2020pretrained} projected to 256 dimensions, plus a learned positional embedding and a 256-dimensional projection of structural features. These structural features consist of the sentence's position within the note, the note's position within the admission, the admission's length in days, the time since the first note and until discharge, how often the same sentence recurs in other notes, its token length, the note's title, and the section header it falls under.

The admission's sentence representations then pass through a two-layer, eight-head Transformer encoder with hidden size 256 and a linear classification head. A sentence's salience can therefore depend on its context, for example whether it is repeated or rendered unimportant by its surrounding course. We fine-tune the final eight layers of BioClinRoBERTa jointly with the rest of the model and train with binary cross-entropy, weighting positive examples by the class ratio of each block.


To isolate the contributions of fine-tuning and of the Transformer encoder, we train two sentence-level ablations that share the same per-sentence representation but score each sentence on its own, without attending to the admission's other sentences. \emph{\omitehr{}-frozen} keeps BioClinRoBERTa frozen and trains only the feature projections and classification head, while \emph{\omitehr{}-fine-tuned} also updates its final eight layers. We repeat all three variants with BioClinicalBERT \citep{alsentzer2019publicly} and BioClinical ModernBERT \citep{sounack2025bioclinical} as the base encoder. Appendix~\ref{apd:model_architecture} provides further architectural and training details.

\subsection{Baselines and Evaluation Protocol}
\label{methods:baselines}
\paragraph{Baselines.}
We compare against three families of baselines. \emph{Structural heuristics} score a sentence with a single scalar (relative position in the record, length in tokens, or repetition count) and test whether salience reduces to document structure. \emph{Lexical models} test whether salience reduces to wording. We use a token-level naive Bayes classifier, with and without length normalization, and logistic-regression and MLP classifiers over TF-IDF sentence vectors. Finally, the \emph{LLM judge} mirrors the per-sentence importance judging of inference-time detectors \citep{bedi2026care}. For each sentence, we prompt GPT-5-mini to estimate the probability that a physician would include its content in the BHC. Each call explicitly supplies up to 20 consecutive source sentences to score, together with shared admission context consisting of the first 40,000 characters of the admission’s notes. The judge always receives each sentence it scores together with neighboring sentences in the same batch, regardless of its position in the record; the 40,000-character limit applies only to the shared admission context. This shared context gives the judge admission-level information to promote a fair comparison to \omitehr{}-contextual.

\begin{table*}[t]
\centering
\footnotesize
\setlength{\tabcolsep}{10pt}
\begin{tabular}{llccc}
\toprule
\textbf{Family} & \textbf{Model} & \textbf{AUROC} & \textbf{AUPRC} & \textbf{R@100} \\
\midrule
\multirow{3}{*}{Heuristic}
& Position   & $0.537\pm.004$ & $0.394\pm.007$ & $0.313\pm.014$ \\
& Length     & $0.593\pm.004$ & $0.452\pm.006$ & $0.360\pm.014$ \\
& Repetition & $0.506\pm.004$ & $0.379\pm.007$ & $0.292\pm.013$ \\
\midrule
\multirow{4}{*}{Lexical}
& Token naive Bayes                     & $0.730\pm.004$ & $0.588\pm.008$ & $0.458\pm.016$ \\
& Token naive Bayes (length-normalized) & $0.740\pm.004$ & $0.567\pm.008$ & $0.448\pm.017$ \\
& TF--IDF + logistic regression         & $0.767\pm.003$ & $0.621\pm.007$ & $0.476\pm.016$ \\
& TF--IDF + MLP                         & $0.730\pm.004$ & $0.574\pm.007$ & $0.451\pm.016$ \\
\midrule
LLM judge & GPT-5-mini (zero-shot) & $0.806\pm.004$ & $0.668\pm.008$ & $0.519\pm.018$ \\
\midrule
\multirow{3}{*}{\omitehr{}-frozen}
& BioClinRoBERTa         & $0.789\pm.003$ & $0.647\pm.007$ & $0.492\pm.017$ \\
& BioClinicalBERT        & $0.780\pm.003$ & $0.635\pm.007$ & $0.485\pm.016$ \\
& BioClinical ModernBERT & $0.784\pm.003$ & $0.639\pm.007$ & $0.488\pm.017$ \\
\midrule
\multirow{3}{*}{\omitehr{}-fine-tuned}
& BioClinRoBERTa         & $0.825\pm.003$ & $0.691\pm.007$ & $0.519\pm.018$ \\
& BioClinicalBERT        & $0.819\pm.003$ & $0.682\pm.007$ & $0.515\pm.018$ \\
& BioClinical ModernBERT & $0.816\pm.003$ & $0.677\pm.007$ & $0.512\pm.018$ \\
\midrule
\multirow{3}{*}{\textbf{\omitehr{}-contextual}}
& \textbf{BioClinRoBERTa} & $\mathbf{0.828\pm.003}$ & $\mathbf{0.697\pm.007}$ & $\mathbf{0.522\pm.018}$ \\
& BioClinicalBERT        & $0.821\pm.003$ & $0.685\pm.007$ & $0.517\pm.018$ \\
& BioClinical ModernBERT & $0.816\pm.004$ & $0.677\pm.007$ & $0.513\pm.018$ \\
\bottomrule
\end{tabular}
\caption{Admission-level salience classification on the held-out test set (750 admissions, over 446K sentences). R@100 is salience recall among the 100 highest-ranked sentences per admission. Each cell reports the point estimate $\pm$ the half-width of the 95\% hierarchical-bootstrap confidence interval. Bold marks the best result.}
\label{tab:salience_full}
\end{table*}

\paragraph{Evaluation.}
\label{methods:evaluation}
All models are evaluated on a held-out test set of 750 admissions containing over 446K sentences, 37\% of which the labeling pipeline marks salient. For each admission we compute AUROC and AUPRC over that admission's sentences alone, then average across admissions. Confidence intervals come from a hierarchical bootstrap over admissions. We test whether the contextual model outperforms the strongest sentence-level model with a paired bootstrap over admissions with 20{,}000 replicates.

\subsection{Omission Detection}
\label{methods:deployment}
OmitEHR can be applied to any system that generates a BHC. At inference time, we apply the same preprocessing used during training and score every source sentence with the trained salience model. We retain the $K$ highest-scoring unique sentences, where $K$ controls the size of the candidate pool. We use $K=100$ throughout our experiments to provide a broad candidate pool for coverage assessment, though the cutoff can be tuned to deployment needs. An LLM judge then checks whether each retained sentence is already covered by the generated summary, using the same inclusion rubric used in salience label generation (Table~\ref{tab:rubric}), now applied to the LLM-generated BHC. As defined in Section~\ref{sec:formulation}, sentences among the \(K\) highest-ranked candidates that receive a coverage label of 0 are surfaced as candidate omissions. Additional details on the omission detection pipeline are provided in Appendix~\ref{apd:omission_surfacing}.



\subsection{Clinician Evaluation of Omissions}
\label{methods:clinician}
To evaluate whether the candidates \omitehr{} surfaces are omissions a clinician would act on in practice, and to compare them with the candidates an LLM pipeline surfaces at the same review budget, we ran a blinded clinician evaluation. We sampled 25 admissions from the test set, generated a BHC for each with Gemini-2.5-pro, and ran both \omitehr{} and the full LLM-judge pipeline.  The pipelines differed in salience scoring: \omitehr{} used its learned salience model, while the full LLM pipeline used GPT-5-mini to score clinical importance; both used Gemini-2.5-pro for coverage assessment with $K=100$. The 10 highest-ranked candidate omissions from each pipeline per admission were reviewed independently by two board-certified clinicians. For each admission, we provided the candidate omission, along with the complete source notes from the admission and the LLM-generated BHC (Figure~\ref{fig:ui_mini}).
Since this was a retrospective annotation study where reviewers had not participated in the patients’ care, each candidate omission was accompanied by a brief LLM-generated rationale derived from the source notes. This rationale situates the sentence in the admission's timeline, cause, and history. It withholds any verdict on inclusion and is not part of the detection pipeline. Prompts are included in Appendix~\ref{apd:llm_prompts}.

\section{Results}
\label{sec:results}

\subsection{Salience Classification}
\label{results:predictions}

Table~\ref{tab:salience_full} reports the AUROC, AUPRC, and recall at 100 (R@100) for salience prediction, computed within each admission and then averaged, so each model is judged on how well it ranks the sentences of a single record. R@100 is the fraction of an admission's salient sentences ranked in its top 100, matching our coverage-labeling budget. Structural heuristics perform near chance, with the best, sentence length, reaching only 0.593 AUROC and 0.360 R@100. This suggests that salience is not reducible to document structure. Lexical models perform markedly better (TF--IDF with logistic regression: 0.767 AUROC, 0.621 AUPRC), suggesting that vocabulary carries substantial signal. The GPT-5-mini judge reaches 0.806 AUROC and 0.668 AUPRC, better than every lexical model and \omitehr{}-frozen.



Fine-tuning the encoder accounts for most of the improvement among the \omitehr{} variants. Tuning the final eight layers of BioClinRoBERTa while still scoring each sentence independently raises performance by 3.6 AUROC points and 4.4 AUPRC points over the frozen encoder and also surpasses the LLM judge when evaluated against the LLM-derived inclusion label. Adding the contextual encoder provides an additional gain of 0.3 AUROC points and 0.6 AUPRC points. This improvement, while modest, is statistically significant under a paired bootstrap $(p<0.0001)$, but it comes at additional computational cost, so the fine-tuned sentence-level model remains a viable lighter-weight option. Across encoders, BioClinRoBERTa is the strongest for every model class, fine-tuning improves every encoder, and context adds a small gain for BioClinRoBERTa and BioClinicalBERT but none for BioClinical ModernBERT. Additional salience results, including evaluation against clinician annotations, are reported in Appendix~\ref{apd:full_results}.

\subsection{Clinician Evaluation}
\label{results:clinician}

\begin{figure*}[t]
  \centering
  \includegraphics[width=0.95\textwidth]{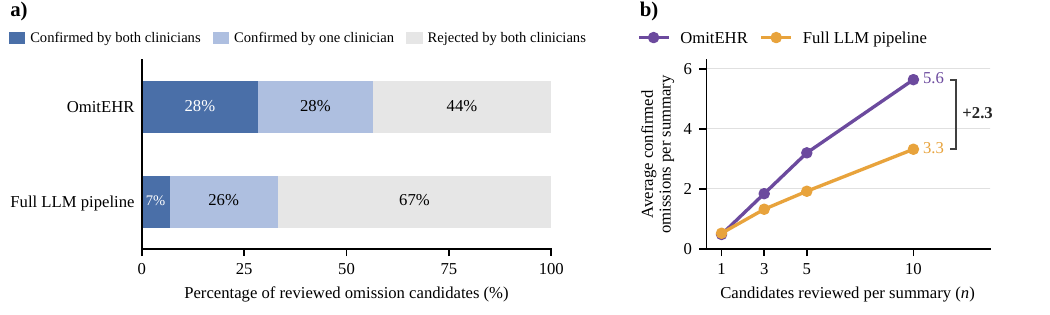}
  \caption{Blinded clinician review of candidate omissions surfaced by OmitEHR and the full LLM pipeline (GPT-5-mini) on 25 LLM-generated BHCs. (a)~Verdicts from two board-certified clinicians on the ten highest-ranked candidates per admission (250 per pipeline), each reviewed independently. (b)~Mean number of omissions per summary confirmed by at least one clinician among the top $n$ ranked candidates. At $n = 10$, OmitEHR surfaced 2.3 more confirmed omissions per summary than the full LLM pipeline. Additional precision results reporting confirmation by both clinicians are reported in Table~\ref{tab:precision_budget}.}
  \label{fig:clinical-review}
\end{figure*}

Two board-certified clinicians independently reviewed the ten highest-ranked candidate omissions from \omitehr{} and the full LLM-judge pipeline on the same 25 LLM-generated BHCs (Figure~\ref{fig:clinical-review}). Of the 250 \omitehr{} candidates reviewed, 141 (56.4\%) were judged to be clinically meaningful omissions by at least one clinician and 71 (28.4\%) by both, compared with 83 (33.2\%) and 17 (6.8\%), respectively, for the full LLM pipeline (Figure~\ref{fig:clinical-review}a). Per summary, \omitehr{} surfaced a mean of 5.64 omissions confirmed by at least one clinician and 2.84 confirmed by both, compared with 3.32 and 0.68 for the full LLM pipeline. \omitehr{} also surfaced at least one omission confirmed by both clinicians in 24 of 25 admissions, compared with 12 of 25 for the full LLM pipeline. The two pipelines performed
similarly when clinicians reviewed only the top-ranked candidate, and \omitehr{}'s advantage widened as the review budget grew (Figure~\ref{fig:clinical-review}b). Precision at each review budget and per-clinician decisions are reported in Appendix~\ref{apd:omission_review}.

\subsection{Cost and Latency}
\label{results:cost}

\begin{figure*}[t]
  \centering
  \includegraphics[width=0.95\textwidth]{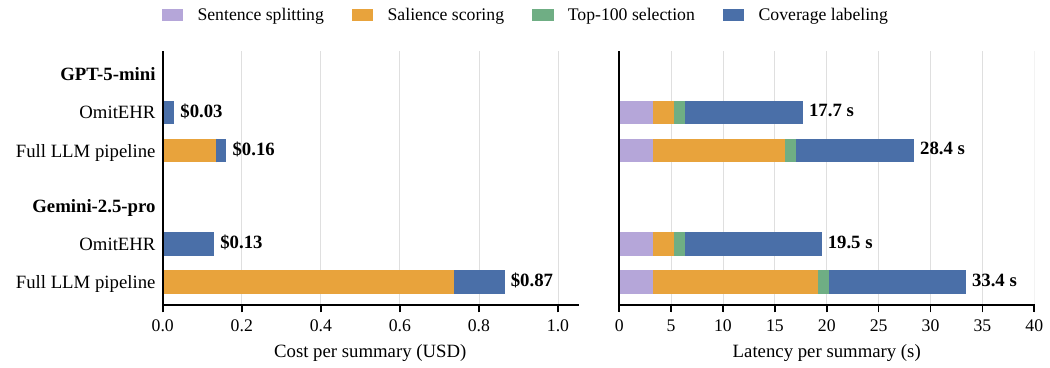}
  \caption{Mean cost and latency of screening one generated BHC, by pipeline stage, across the 25 evaluation admissions. Each row uses the specified model for all LLM calls. Within each row, \omitehr{} and the full LLM pipeline differ only in salience scoring: \omitehr{} uses its learned salience model, whereas the full LLM pipeline scores every source sentence with the LLM. \omitehr{}'s salience step incurs approximately \$0.002 in GPU cost and is too small to be visible. Latencies are wall-clock times with up to 16 concurrent API calls. Details in Appendix~\ref{apd:token_costs}.}
  \label{fig:cost}
\end{figure*}

Averaged over the 25 evaluation admissions, screening one generated BHC with \omitehr{} takes 17.7\,s for \$0.03 with GPT-5-mini for coverage labeling and 19.5\,s for \$0.13 with Gemini-2.5-pro. We compare \omitehr{} against a full LLM pipeline that uses the same model for both salience and coverage (Figure~\ref{fig:cost}).

The first three stages of \omitehr{}---sentence splitting, salience scoring, and top-$K$ selection with $K=100$---take 6.3\,s and require no LLM API calls. Coverage labeling is the only LLM-dependent stage.

The full LLM pipeline differs only in how salience is assessed: rather than using \omitehr{}'s learned salience model, it asks an LLM to score every source sentence in the admission. LLM-based salience assessment alone consumes 10 to 11 times as many input tokens as \omitehr{}'s entire pipeline and takes 12.7--15.9\,s, compared with 2.0\,s for \omitehr{}'s salience model. End to end, \omitehr{} is 5.6 to 6.7 times cheaper and 1.6 to 1.7 times faster, and its LLM cost is capped at $K$ whereas the full pipeline's grows with the number of sentences. Additional comparisons of cost and latency are located in Appendix~\ref{apd:token_costs}.

\section{Discussion}
\label{sec:discussion}

This work demonstrates that reference-free omission detection need not infer clinical importance from scratch at inference time. Instead, salience can be learned from historical clinician-written summaries and used to identify important source content. Using this supervision, a compact contextual salience model outperforms  structural, lexical, and LLM-judge baselines at predicting which source content belongs in a BHC. \omitehr{} combines this learned notion of salience with coverage assessment to surface clinically important information omitted from an LLM-generated summary.


Learning salience also changes the computational structure of omission detection. Inference-time LLM judges must repeatedly inspect the source record to determine what matters, causing salience-assessment cost to grow with record length. \omitehr{} instead predicts salience with a learned model and reserves LLM-based coverage assessment for a fixed set of top-$K$ candidates. As a result, the expensive LLM component is bounded by $K$ rather than by the number of source sentences. 
This distinction matters for real-time clinical use: an omission detector must add minimal latency if it is to screen summaries without slowing clinical workflows. For long longitudinal records, reading the full chart can otherwise dominate both latency and cost.

The clinician evaluation suggests that clinically meaningful omissions remain present in LLM-generated summaries. \omitehr{} surfaced a mean of 5.64 omissions per summary confirmed by at least one clinician, compared with 3.32 for the full LLM-judge pipeline. It also surfaced at least one omission confirmed by both clinicians in 24 of 25 admissions, compared with 12 of 25 for the baseline. These results suggest that learned salience can more effectively direct clinicians to clinically important missing information.



The \omitehr{} recipe can extend to any summary type with a history of clinician-written references, such as specialist consult notes or oncology progress notes, and conditioning the salience model on the querying provider, specialty, and clinical question is a direct next step. Beyond auditing, the same framework could support summary generation itself. Surfaced candidates can be fed back to the summarizer for revision, and predicted salience can guide generation directly, as SPEER does with entity-level labels \citep{adams2024speer}.

\paragraph{Limitations.}
\label{sec:limitations}
Our study has several limitations that motivate future work. First, we focus only on brief hospital course generation from internal medicine admissions at a single academic medical center. Future work should evaluate generalization to other sites, services, and summary types. Second, our salience evaluation relies on silver-standard labels that show moderate agreement with clinician annotations ($\kappa = 0.41$ to $0.45$). Accordingly, the results in Table~\ref{tab:salience_full} measure agreement with an LLM-derived target rather than with clinical judgment directly. These labels allow for evaluation at scale but inherit limitations from both the LLM-based labeling procedure and the clinician-written summaries themselves, which may not capture all clinically important information. We therefore complement the salience evaluation with a blinded clinician evaluation to provide an independent downstream assessment of whether the surfaced candidates are clinically meaningful. Third, our omission evaluation measures precision among the surfaced candidates, not recall over all clinically important omissions. Measuring recall would require clinicians to review entire records, about 580 sentences per admission, which was beyond our annotation budget. The salience Recall@$K$ and candidate-count analyses capture the trade-off between computational cost and salience coverage but do not establish omission recall, which remains an experiment for future work. Fourth, our clinician evaluation was limited to 25 admissions and two reviewers, neither of whom had prior knowledge of the cases they reviewed. LLM-generated rationales were provided to contextualize each candidate omission but may have influenced judgments despite instruction to withhold an explicit omission verdict. Additionally, the absence of covered or low-salience control candidates in our clinician evaluation prevents estimation of reviewers' false-positive rate. Future work should expand clinician evaluation to familiar cases, include blinded control candidates, and compare judgments with and without rationales.




\bibliography{ref}

\appendix
\section{Dataset and Label Distribution}\label{apd:dataset}

The 5,000 admissions are split by patient into training, development, and held-out test sets, so no patient contributes to more than one split. The training, development and test sets contain 3500, 750 and 750 admissions respectively. Table~\ref{tab:entailment-label-distribution} reports the distribution of silver inclusion labels in each split. Class proportions are stable across splits, with 36--38\% of sentences salient under the binary target (labels 1 and 2).

\begin{table}[H]
  \centering
  \small
  \setlength{\tabcolsep}{4pt}
  \resizebox{\columnwidth}{!}{%
    \begin{tabular}{lrrrr}
      \toprule
      Split & Sentences & 0: Not related & 1: Indirectly related & 2: Directly included \\
      \midrule
      Train & 2{,}054{,}630 & 1{,}312{,}014 (63.9\%) & 620{,}859 (30.2\%) & 121{,}757 (5.9\%) \\
      Dev   & 397{,}998     & 248{,}090 (62.3\%)     & 125{,}557 (31.5\%) & 24{,}351 (6.1\%) \\
      Test  & 446{,}467     & 282{,}259 (63.2\%)     & 138{,}679 (31.1\%) & 25{,}529 (5.7\%) \\
      \bottomrule
    \end{tabular}%
  }
  \caption{Silver inclusion-label distribution by split. Values are sentence counts with within-split percentages in parentheses.}
  \label{tab:entailment-label-distribution}
\end{table}

\section{Entailment Label Validation}\label{apd:entailment}

\begin{figure*}[t]
  \centering
  \includegraphics[width=\textwidth]{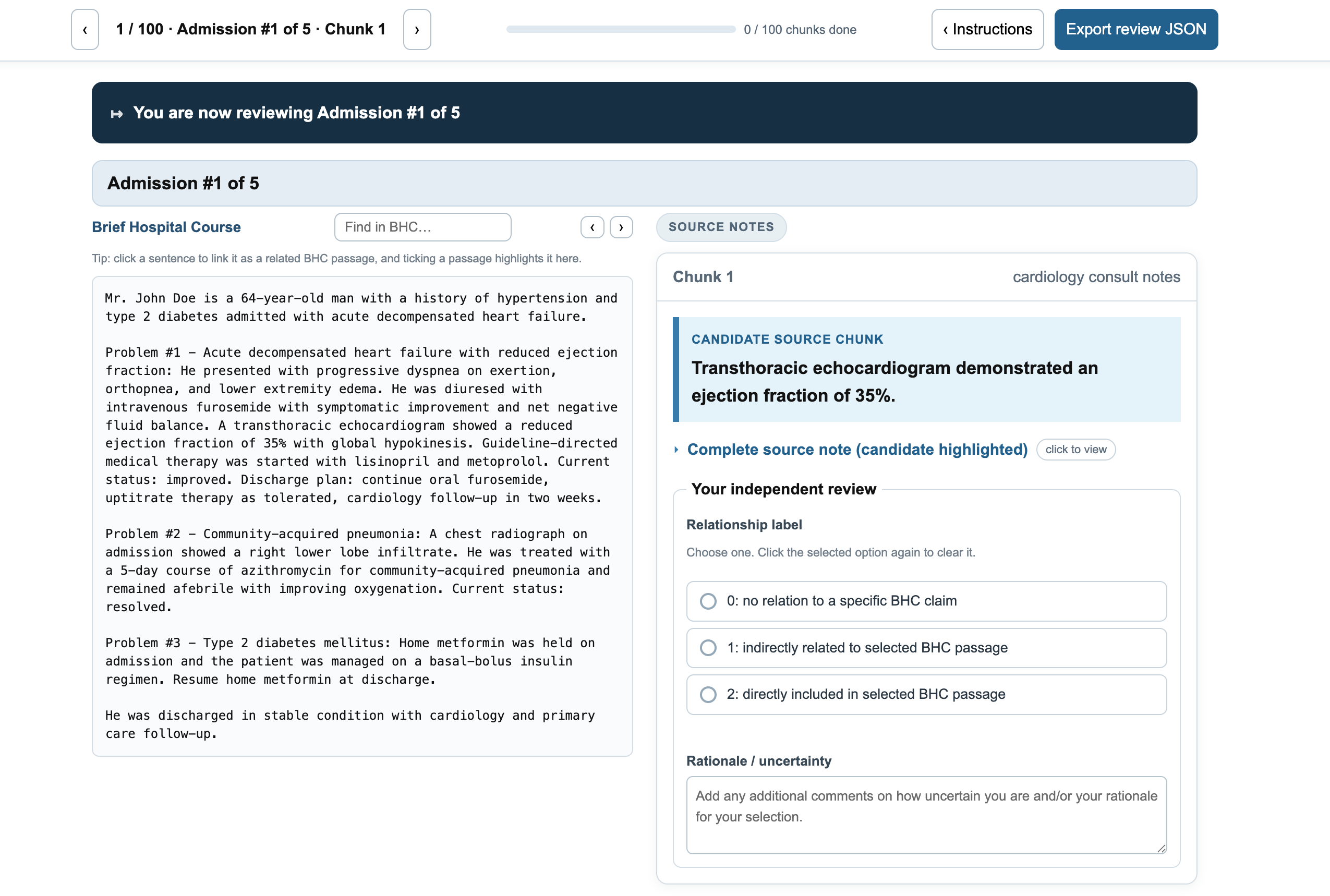}
  \caption{Interface used by clinicians to review source chunks and assign
    entailment labels relative to the corresponding clinician-written BHC.}
  \label{fig:entailment_review_interface}
\end{figure*}

To assess the entailment labeling quality of the LLM Judge, two internal medicine clinicians independently annotated the same 300 sentences sampled from 15 distinct admissions. Both were blinded to the LLM labels and to each other's annotations, and each sentence was presented alongside the full source note and BHC. An example screen of the annotation platform is shown in Figure~\ref{fig:entailment_review_interface}.

The two clinicians agreed with each other on 81.0\% of sentences (linearly weighted $\kappa = 0.647$, 95\% CI 0.560 to 0.728). We treat this as the practical ceiling for the task, since a labeler cannot be expected to agree with either clinician more closely than the clinicians agree with each other. The clinicians differ on 57 of 300 sentences, most often at the boundary between labels~0 and~1 (33 of the 57). GPT-5-mini reaches 70.7\% and 68.7\% exact agreement ($\kappa = 0.454$ and $0.414$) against the two clinicians respectively (Table~\ref{tab:agreement}).

\begin{table}[!ht]
  \centering
  \small
  \setlength{\tabcolsep}{4pt}
  \begin{tabular}{@{}lcc@{}}
    \toprule
    Raters & Exact agr. & Wtd.\ $\kappa$ \\
    \midrule
    Clinician~1 vs.\ 2          & 81.0\% & 0.647 \\
    Clinician~1 vs.\ GPT-5-mini & 70.7\% & 0.454 \\
    Clinician~2 vs.\ GPT-5-mini & 68.7\% & 0.414 \\
    \bottomrule
  \end{tabular}
  \caption{Pairwise annotation agreement on the 300-sentence blinded review set.
    Weighted $\kappa$ uses linear weights across the ordered three-class label
    scheme.}
  \label{tab:agreement}
\end{table}


Out of the 300 reviewed sentences, the labels for 185 were agreed upon by all three parties (Clinician~1, Clinician~2, and GPT-5-mini). A third annotator independently reviewed the remaining 115 disputed sentences in a blinded review. Among the 58 cases in which Clinicians~1 and~2 agreed with each other but disagreed with GPT-5-mini, the adjudicator agreed with the LLM in 26 cases (44.8\%; Table~\ref{tab:c3_adjudication}). This near-even split suggests that the remaining disagreements are mostly sentences that sit between two labels, where annotators themselves differ, rather than systematic errors by the LLM.

\begin{table}[t]
  \centering
  \small
  \setlength{\tabcolsep}{3pt}
  \begin{tabular}{@{}
    >{\raggedright\arraybackslash}p{0.52\columnwidth}
    >{\centering\arraybackslash}p{0.07\columnwidth}
    >{\centering\arraybackslash}p{0.25\columnwidth}
    @{}}
    \toprule
    Initial disagreement pattern & $n$ & Adjudicator agrees with LLM \\
    \midrule
    R1 and R2 agree against LLM
      & 58 & 26 (44.8\%) \\
    R1 and R2 split, one agrees with LLM
      & 48 & 31 (64.6\%) \\
    R1 and R2 split, neither agrees with LLM
      & 9 & 4 (44.4\%) \\
    \midrule
    All initially disputed sentences
      & 115 & 61 (53.0\%) \\
    \bottomrule
  \end{tabular}
  \caption{Adjudicator review of the 115 sentences for which at least one
  clinician disagreed with GPT-5-mini. R1 and R2 denote the two clinicians.}
  \label{tab:c3_adjudication}
\end{table}

\section{Salience Model and Training Details}\label{apd:model_architecture}

\begin{figure}[t]
  \centering
  \includegraphics[width=0.9\columnwidth]{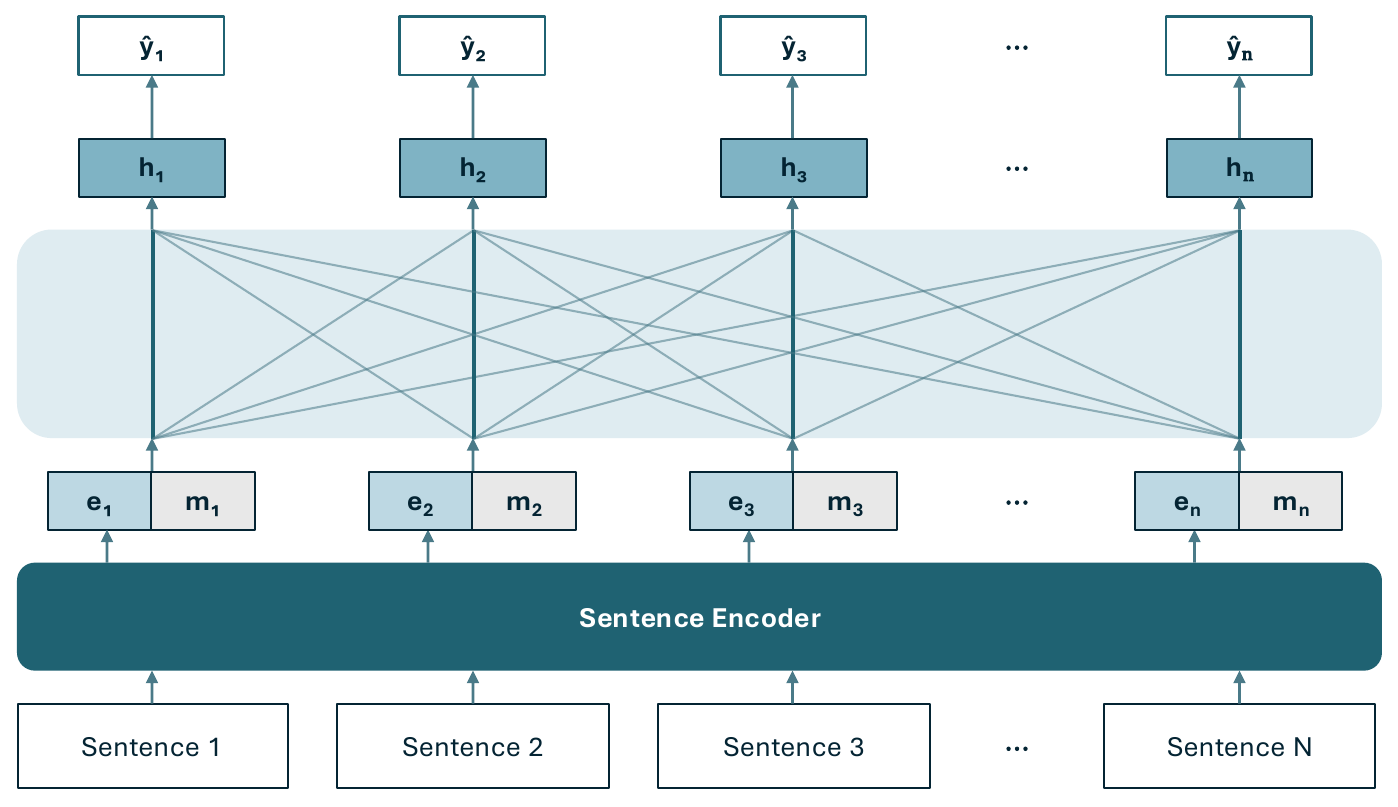}
  \caption{The contextual salience model. Each sentence embedding $e_i$ is combined with projected structural features $m_i$ and a position embedding, contextualized by an admission-level Transformer encoder, and scored by a per-sentence head.}
  \label{fig:architecture}
\end{figure}

\paragraph{Architecture.} Figure~\ref{fig:architecture} shows the architecture of OmitEHR-contextual. Each source sentence is encoded independently with RoBERTa-large-PM-M3-Voc \citep{lewis2020pretrained}. The final-layer token representations are mean-pooled into one sentence embedding and projected to 256 dimensions. A second 256-dimensional vector is projected from the structural features and added to it, along with a learned sentence-position embedding. The structural features are relative position within the admission and within the note, admission length, time since the first note and until discharge, within-admission repetition count, token length, and learned embeddings of the note title (hashed into 128 buckets) and of a rule-based section category (9 categories). The resulting sequence passes through a two-layer, eight-head Transformer encoder with hidden dimension 256, GELU activations, and dropout 0.1. Admissions are divided into contiguous blocks of at most 1,024 sentences that retain source order. A layer-normalized linear head produces one logit per sentence, converted by a sigmoid into a salience probability.

\paragraph{Training.}
The model is trained with class-weighted binary cross-entropy, treating silver labels 1 and 2 as salient and label 0 as non-salient, and weighting positive examples by the negative-to-positive ratio of the training block. The final eight encoder layers are fine-tuned jointly with the projections, contextual encoder, and head, while earlier encoder layers are frozen. The final checkpoint is selected by admission-level AUPRC on the development set. The sentence-level variants (\omitehr{}-frozen and \omitehr{}-fine-tuned) use the same sentence representation, structural features, and head, but omit the contextual encoder and score each sentence independently. We use AdamW with weight decay 0.01 and gradient clipping at 1.0, with a learning rate of $1\times10^{-5}$ for the fine-tuned encoder layers and $3\times10^{-4}$ for the contextual encoder, projections, and head. Training runs for three epochs on a single NVIDIA H100 GPU with mixed precision and gradient checkpointing.
\section{Additional Salience Results}\label{apd:full_results}

This appendix supplements Table~\ref{tab:salience_full} in three ways. First, we report recall at $K$ for every model. Second, we repeat salience training and evaluation with direct inclusion as the only positive class, meaning label~2 of the rubric in Table~\ref{tab:rubric}, content stated directly in the BHC, rather than labels~1 and~2 together. Third, we rescore \omitehr{}-contextual and the LLM judge on the 300 test sentences the clinicians annotated in Appendix~\ref{apd:entailment} to check whether a ranking learned from LLM labels holds up against clinician judgment.


\paragraph{Salience recall.}
Table~\ref{tab:salience_recall} reports the recall of the salience model at $K$ (R@$K$), the fraction of an admission's salient sentences ranked among its top $K$, averaged over admissions, for $K \in \{10, 50, 100, 200, 400\}$. $K=100$ corresponds to the number of candidate sentences passed to coverage labeling in the main pipeline (Section~\ref{results:cost}). At every cutoff, the fine-tuned and contextual \omitehr{} models outperform all heuristic and lexical baselines, and contextual BioClinRoBERTa has the highest or tied-highest recall.

\begin{table*}[t]
\centering
\footnotesize
\setlength{\tabcolsep}{3pt}
\resizebox{\textwidth}{!}{%
\begin{tabular}{llccccc}
\toprule
\textbf{Family} & \textbf{Model} & R@10 & R@50 & R@100 & R@200 & R@400 \\
\midrule
\multirow{3}{*}{Heuristic}
& Position   & $0.031\pm.002$ & $0.158\pm.009$ & $0.313\pm.014$ & $0.567\pm.020$ & $0.832\pm.018$ \\
& Length     & $0.047\pm.003$ & $0.200\pm.009$ & $0.360\pm.014$ & $0.606\pm.018$ & $0.847\pm.017$ \\
& Repetition & $0.030\pm.002$ & $0.148\pm.008$ & $0.292\pm.013$ & $0.547\pm.019$ & $0.822\pm.018$ \\
\midrule
\multirow{4}{*}{Lexical}
& Token naive Bayes                     & $0.058\pm.003$ & $0.267\pm.011$ & $0.458\pm.016$ & $0.696\pm.017$ & $0.888\pm.015$ \\
& Token naive Bayes (length-normalized) & $0.051\pm.003$ & $0.247\pm.011$ & $0.448\pm.017$ & $0.703\pm.019$ & $0.896\pm.015$ \\
& TF--IDF + logistic regression         & $0.065\pm.003$ & $0.275\pm.012$ & $0.476\pm.016$ & $0.720\pm.018$ & $0.901\pm.014$ \\
& TF--IDF + MLP                         & $0.057\pm.003$ & $0.254\pm.011$ & $0.451\pm.016$ & $0.696\pm.018$ & $0.889\pm.015$ \\
\midrule
LLM judge & GPT-5-mini (zero-shot) & $\mathbf{0.071\pm.004}$ & $0.307\pm.013$ & $0.519\pm.018$ & $0.748\pm.018$ & $0.905\pm.013$ \\
\midrule
\multirow{3}{*}{\omitehr{}-frozen}
& BioClinRoBERTa         & $0.066\pm.003$ & $0.286\pm.012$ & $0.492\pm.017$ & $0.736\pm.018$ & $0.908\pm.014$ \\
& BioClinicalBERT        & $0.066\pm.003$ & $0.280\pm.012$ & $0.485\pm.016$ & $0.729\pm.018$ & $0.905\pm.014$ \\
& BioClinical ModernBERT & $0.064\pm.003$ & $0.283\pm.012$ & $0.488\pm.017$ & $0.733\pm.018$ & $0.906\pm.014$ \\
\midrule
\multirow{3}{*}{\omitehr{}-fine-tuned}
& BioClinRoBERTa         & $\mathbf{0.071\pm.004}$ & $0.305\pm.013$ & $0.519\pm.018$ & $0.761\pm.018$ & $0.918\pm.013$ \\
& BioClinicalBERT        & $0.069\pm.004$ & $0.301\pm.013$ & $0.515\pm.018$ & $0.757\pm.018$ & $0.916\pm.014$ \\
& BioClinical ModernBERT & $0.069\pm.004$ & $0.298\pm.013$ & $0.512\pm.018$ & $0.754\pm.018$ & $0.915\pm.014$ \\
\midrule
\multirow{3}{*}{\textbf{\omitehr{}-contextual}}
& \textbf{BioClinRoBERTa} & $\mathbf{0.071\pm.004}$ & $\mathbf{0.308\pm.013}$ & $\mathbf{0.522\pm.018}$ & $\mathbf{0.763\pm.017}$ & $\mathbf{0.919\pm.013}$ \\
& BioClinicalBERT        & $0.069\pm.004$ & $0.302\pm.013$ & $0.517\pm.018$ & $0.758\pm.018$ & $0.917\pm.014$ \\
& BioClinical ModernBERT & $0.069\pm.004$ & $0.298\pm.013$ & $0.513\pm.018$ & $0.755\pm.018$ & $0.916\pm.013$ \\
\bottomrule
\end{tabular}}
\caption{Admission-level salience recall at $K$ on the held-out test set (750 admissions). R@$K$ is the fraction of an admission's salient sentences ranked among its top $K$, averaged over admissions. Each cell reports the point estimate $\pm$ the half-width of the 95\% hierarchical-bootstrap confidence interval. Bold marks the best result in each column.}
\label{tab:salience_recall}
\end{table*}


\paragraph{Training on direct inclusion labels.}
Table~\ref{tab:salience_full} treats labels~1 and~2 (see Table~\ref{tab:rubric}) as salient, so its positive class is cut at the boundary between labels~0 and~1, which is where the clinicians disagree most often (Appendix~\ref{apd:entailment}). We therefore repeated the experiment with label~2 alone, content stated directly in the BHC, as both the training target and the evaluation target. We retrained the lexical and \omitehr{} models on label-2 training labels and scored every model, including the heuristics and the GPT-5-mini judge, against label-2 test labels. Table~\ref{tab:salience_label2} reports the best heuristic and lexical baselines, the judge, and the two trained \omitehr{} variants. On the test set, 5.7\% of sentences have label~2. Contextual BioClinRoBERTa reaches 0.886 AUROC and 0.356 AUPRC, and its top 100 sentences capture 78.5\% of an admission's label-2 sentences. The GPT-5-mini judge falls below the lexical baseline on this target.

\begin{table*}[t]
\centering
\footnotesize
\setlength{\tabcolsep}{4pt}
\begin{tabular}{lccccccc}
\toprule
\textbf{Model} & AUROC & AUPRC & R@10 & R@50 & R@100 & R@200  \\
\midrule
  Position                      & $0.589\pm.010$ & $0.087\pm.005$ & $0.025\pm.006$ & $0.172\pm.016$ & $0.366\pm.022$ & $0.629\pm.023$ \\
  TF--IDF + logistic regression & $0.807\pm.007$ & $0.246\pm.011$ & $0.145\pm.011$ & $0.460\pm.019$ & $0.657\pm.019$ & $0.844\pm.016$  \\
  GPT-5-mini judge (zero-shot)  & $0.749\pm.007$ & $0.149\pm.007$ & $0.080\pm.009$ & $0.365\pm.020$ & $0.577\pm.021$ & $0.775\pm.018$ \\
  \omitehr{}-fine-tuned         & $0.882\pm.005$ & $0.347\pm.013$ & $0.216\pm.015$ & $0.590\pm.019$ & $0.777\pm.017$ & $0.910\pm.012$ \\
  \textbf{\omitehr{}-contextual} & $\mathbf{0.886\pm.005}$ & $\mathbf{0.356\pm.014}$ & $\mathbf{0.223\pm.015}$ & $\mathbf{0.592\pm.019}$ & $\mathbf{0.785\pm.017}$ & $\mathbf{0.913\pm.012}$ \\
  \bottomrule

\end{tabular}
\caption{Admission-level salience classification and recall at $K$ with label~2 as the only positive class, on the held-out test set (750 admissions, 5.7\% positive). \omitehr{} models use BioClinRoBERTa. Position and TF--IDF + logistic regression are the best heuristic and lexical baselines. R@$K$ is defined as in Table~\ref{tab:salience_recall}. Each cell reports the point estimate $\pm$ the half-width of the 95\% hierarchical-bootstrap confidence interval. Bold marks the best result.}
\label{tab:salience_label2}
\end{table*}

\paragraph{Evaluation against clinician labels.}
Table~\ref{tab:salience_full} evaluates models against GPT-5-mini's labels, which show only moderate agreement with clinician annotations. To assess alignment with physician judgment, Table~\ref{tab:salience_gold} rescores \omitehr{}-contextual and the LLM judge on the 300 clinician-annotated test sentences described in Appendix~\ref{apd:entailment}, against both the original GPT-5-mini labels (silver standard) and the clinician majority labels (gold standard). Against gold labels the original \omitehr{} model reaches 0.820 AUROC and 0.702 AUPRC, close to its silver-label results on the same sentences (0.842 and 0.717 respectively). The LLM Judge's point estimates are higher against the gold labels, consistent with silver-label evaluation favoring the model trained on those labels. However, neither AUROC nor AUPRC differs significantly between the two models, leaving insufficient evidence to distinguish their ranking performance. With direct inclusion as the target, the model trained on label~2 only outperforms the judge reprompted for that target against both silver and gold labels. The difference is significant for AUROC against silver labels (0.876 versus 0.692, $p = 0.031$) and for AUPRC against gold labels (0.502 versus 0.318, $p = 0.018$).


\begin{table*}[t]
\centering
\footnotesize
\setlength{\tabcolsep}{5pt}
\begin{tabular}{llcccccc}
\toprule
& & \multicolumn{3}{c}{\textbf{AUROC}} & \multicolumn{3}{c}{\textbf{AUPRC}} \\
\cmidrule(lr){3-5}\cmidrule(lr){6-8}
\textbf{Training target} & \textbf{Evaluation labels} & \omitehr{} & LLM judge & $p$ & \omitehr{} & LLM judge & $p$ \\
\midrule
\multirow{2}{*}{Labels 1 or 2}
& Silver (GPT-5-mini) & 0.842 & 0.819 & 0.313 & 0.717 & 0.699 & 0.621 \\
& Gold (clinicians)    & 0.820 & 0.857 & 0.137 & 0.702 & 0.764 & 0.203 \\
\midrule
\multirow{2}{*}{Label 2 only}
& Silver (GPT-5-mini) & 0.876 & 0.692 & 0.031 & 0.478 & 0.330 & 0.078 \\
& Gold (clinicians)    & 0.846 & 0.691 & 0.095 & 0.502 & 0.318 & 0.018 \\
\bottomrule
\end{tabular}
\caption{Admission-level salience ranking of \omitehr{}-contextual and the LLM judge on the 300 clinician-annotated sentences of Appendix~\ref{apd:entailment}, scored against GPT-5-mini labels (silver) and clinician majority labels (gold) with the positive class matching the training target. The gold label-2 evaluation uses the 292 sentences with a resolved majority. In the second block the judge is reprompted for direct inclusion. $p$-values are from paired admission-level sign-flip tests.}
\label{tab:salience_gold}
\end{table*}

\section{Omission Candidate Identification}\label{apd:omission_surfacing}

At inference, \omitehr{} ranks source sentences by predicted salience and sends the top \(K\) to the coverage judge, as defined in Section~\ref{sec:formulation}. Here, \(K\) is a configurable inference-time budget controlling the number of sentences evaluated for coverage. For our evaluation, we sampled 25 admissions from the test set and used Gemini 2.5 Pro to generate a brief hospital course (BHC) for each admission. Since notes within an admission often repeat sentences verbatim, we deduplicate identical sentences before ranking so that the top \(K\) candidates are unique. We selected \(K=100\) for all admissions to provide a sufficiently large candidate pool for the subsequent coverage assessment. This yielded 2,500 source sentences for coverage evaluation. Gemini-2.5-pro was used to label each sentence as directly covered, indirectly related, or not covered by the generated BHC using Prompt~C in Appendix~\ref{apd:llm_prompts}. As shown in Table~\ref{tab:bhc-coverage}, the coverage model classified 49.9\% as directly included, 25.1\% as indirectly related, and 25.0\% as not represented in the generated BHC. Sentences judged not covered are treated as candidate omissions and ranked by salience.

\begin{table}[H]
      \centering
      \small
      \begin{tabular}{lr}
      \toprule
      Coverage outcome & $n$ (\%) \\
      \midrule
      Directly included evidence & 1,248 (49.9\%) \\
      Indirectly related & 628 (25.1\%) \\
      Not related (omitted content) & 624 (25.0\%) \\
      \midrule
      Total & 2,500 (100.0\%) \\
      \bottomrule
      \end{tabular}
      \caption{Coverage of the top 100 deduplicated salient source sentences per admission in the review set ($n=25$ admissions; 2,500
      sentences).}
      \label{tab:bhc-coverage}
  \end{table}

\section{Clinician Omission Review Details}\label{apd:omission_review}

For each of the 25 admissions, the two clinician annotators were shown the 10 highest-ranked candidate omissions from both OmitEHR and full LLM pipeline in separate documents, the complete set of notes from the admission, and the generated BHC. Each candidate was additionally accompanied by an LLM-generated rationale (Prompt~D in Appendix~\ref{apd:llm_prompts}) describing what event the sentence refers to, when it occurred, why it happened, and relevant history. This rationale was provided purely as supplementary material to assist the clinicians in familiarizing themselves with the care trajectories of these patients who were not under their care, and it withheld any judgment about whether the sentence should have been included. For each candidate sentence, the clinicians first decided whether the candidate was a true, clinically meaningful omission and, for rejected candidates, whether the information was actually covered by the BHC or not important enough to include.

Table~\ref{tab:omission-review-agreement} summarizes paired clinician decisions on the 250 candidates from each pipeline. For \omitehr{}, the clinicians agreed on 180 candidates, both confirming 71 as true omissions and rejecting 109 (72.0\% agreement; Cohen's $\kappa=0.43$). The remaining 70 candidates were confirmed by only one clinician, yielding 141 candidates (56.4\%) confirmed by at least one clinician. For the full LLM pipeline, both clinicians confirmed 17 candidates and rejected 167 (73.6\% agreement; Cohen's $\kappa=0.18$), with 83 candidates (33.2\%) confirmed by at least one clinician.  Among the 109 \omitehr{} candidates rejected by both clinicians, they agreed on the reason for rejection in 68 instances (62.4\%), including 59 salience errors and 9 coverage errors.



\begin{table}[t]
\centering
\footnotesize
\setlength{\tabcolsep}{4pt}
\begin{tabular}{@{}llrr@{}}
\toprule
Clinician A & Clinician B
& \textbf{\omitehr{}} & \textbf{LLM pipeline} \\
\midrule
True omission & True omission & 71 (28.4)  & 17 (6.8) \\
Not omission  & Not omission  & 109 (43.6) & 167 (66.8) \\
True omission & Not omission  & 46 (18.4)  & 42 (16.8) \\
Not omission  & True omission & 24 (9.6)   & 24 (9.6) \\
\midrule
\multicolumn{2}{@{}l}{Confirmed by either clinician}
& 141 (56.4) & 83 (33.2) \\
\multicolumn{2}{@{}l}{Raw agreement}
& 180 (72.0) & 184 (73.6) \\
\multicolumn{2}{@{}l}{Cohen's $\kappa$}
& 0.43 & 0.18 \\
\bottomrule
\end{tabular}
\caption{Paired clinician decisions on 250 candidates
per pipeline across the same 25 admissions.
Entries report counts (percentages), except for
Cohen's $\kappa$. The one missing clinician judgment
for the LLM pipeline was conservatively treated as
a not-omission decision.}
\label{tab:omission-review-agreement}
\end{table}

\paragraph{Precision by review budget.}
Holding the coverage-assessment budget fixed at $K=100$, we evaluated clinician-confirmed precision among the top $n \in \{1,3,5,10\}$ ranked candidate omissions per admission
across 25 admissions. We use two confirmation criteria. Under the either-clinician criterion, a candidate counts as a confirmed omission if at least one of the two clinicians judged it to be one. Under the both-clinician criterion, both clinicians must have judged it to be one. Under the either-clinician criterion, the full LLM-judge pipeline achieved a Precision@10 of 33.2\% (83/250), compared with 56.4\% (141/250) for \omitehr{}, an absolute improvement of 23.2 percentage points. \omitehr{} also achieved higher precision at top-3 (61.3\% vs.\ 44.0\%) and top-5 (64.0\% vs.\ 38.4\%). Under the stricter both-clinician criterion, Precision@10 was 28.4\% (71/250) for \omitehr{} and 6.8\% (17/250) for the full LLM-judge pipeline. Table~\ref{tab:precision_budget} reports all cutoffs under both criteria.

\begin{table}[t]
\centering
\footnotesize
\setlength{\tabcolsep}{4pt}
\begin{tabular}{@{}lcrr@{}}
\toprule
\textbf{Confirmation} & $\boldsymbol{n}$
& \textbf{\omitehr{}} & \textbf{LLM pipeline} \\
\midrule
Either clinician & 1  & 48.0\% & 52.0\% \\
                 & 3  & 61.3\% & 44.0\% \\
                 & 5  & 64.0\% & 38.4\% \\
                 & 10 & 56.4\% & 33.2\% \\
\midrule
Both clinicians  & 1  & 20.0\% & 16.0\% \\
                 & 3  & 29.3\% & 8.0\%  \\
                 & 5  & 31.2\% & 8.0\%  \\
                 & 10 & 28.4\% & 6.8\%  \\
\bottomrule
\end{tabular}
\caption{Clinician-confirmed Precision@$n$ across 25 admissions, with $K=100$ source sentences checked for coverage. Either-clinician confirmation requires at least one clinician to confirm an omission; both clinician confirmation requires both.}
\label{tab:precision_budget}
\end{table}

\paragraph{Rejection reasons.}
Table~\ref{tab:rejection_reasons} summarizes reasons for rejecting candidates under each pipeline. It includes all candidates rejected by at least one clinician. For both pipelines, clinicians cited insufficient clinical importance more often than existing BHC coverage, suggesting that salience errors were the more common source of rejected candidates.

\begin{table}[t]
\centering
\footnotesize
\setlength{\tabcolsep}{4pt}
\begin{tabular}{@{}lcc@{}}
\toprule
& \textbf{\omitehr{}} & \begin{tabular}[b]{@{}c@{}}\textbf{Full LLM}\\\textbf{pipeline}\end{tabular} \\
\midrule
Rejected by either clinician & 179 & 233 \\
\addlinespace
\multicolumn{3}{@{}l}{\emph{Rejection reasons, $n$ (\% of rejected)}} \\
\quad Clinically unimportant & 146 (81.6) & 185 (79.4) \\
\quad Already covered by BHC & \phantom{0}66 (36.9) & \phantom{0}98 (42.1) \\
\quad Unspecified            & \phantom{00}8 (\phantom{}4.5) & \phantom{0}35 (15.0) \\
\bottomrule
\end{tabular}
\caption{Rejection reasons for candidates rejected by at least one clinician across 25 admissions. Percentages use each pipeline's rejected candidates as the denominator. A candidate counts toward every reason cited by either clinician, so reasons are not mutually exclusive. \emph{Unspecified}: rejected without a reason given.}
\label{tab:rejection_reasons}
\end{table}
\section{Cost and Latency Details}
\label{apd:token_costs}

The estimated one-time cost of offline labeling all 5,000 admissions with GPT-5-mini was \$678.14, or approximately \$0.14 per admission (Table~\ref{tab:token-offline}). All inference-time measurements were taken over the 25 admissions used in the clinician evaluation. Sentence splitting ran on CPU, and salience classification and top-100 selection ran on a single NVIDIA H100 GPU (12 CPU cores, 96\,GB of RAM), with timings excluding model initialization. Coverage labeling sent the top-100 candidates in batches of up to 10, grouped by source note, resulting in about 26 calls per admission. LLM salience scoring supplied up to 20 consecutive source sentences per call, together with shared admission context consisting of the first 40,000 characters of the admission's notes, resulting in about 42 calls per admission. Both processes issued up to 16 requests concurrently within an admission, so latencies are wall-clock times rather than serialized per-request times. Every LLM step was run with both GPT-5-mini and Gemini-2.5-pro (Figure~\ref{fig:cost}). GPT-5-mini salience predictions with Gemini-2.5-pro coverage is the configuration reviewed in the clinician evaluation.

Table~\ref{tab:token-inference} gives per-admission means and Table~\ref{tab:latency} the spread of latencies. Costs are computed from recorded token counts at list prices, \$0.25 per million input tokens and \$2.00 per million output tokens for GPT-5-mini and \$1.25 and \$10.00 for Gemini-2.5-pro. Coverage cost per admission ranged from \$0.020 to \$0.035 with GPT-5-mini (median \$0.027) and from \$0.097 to \$0.166 with Gemini-2.5-pro (median \$0.127), driven by input tokens that ranged from about 25K to 74K depending on the length of the candidates' source notes. The GPU-time estimate for \omitehr{} assumes \$3 per H100-hour over the 49.7 GPU-seconds of salience classification across the 25 admissions.

\begin{table}[H]
\centering
\small
\resizebox{\columnwidth}{!}{%
\begin{tabular}{@{}lrrr@{}}
\toprule
Split & Input tokens & Output tokens & Total cost (USD) \\
\midrule
Train & 734.5M & 114.4M & \$412.37 \\
Dev   & 233.4M &  30.4M & \$119.20 \\
Test  & 303.8M &  35.3M & \$146.57 \\
\midrule
Total & 1{,}271.8M & 180.1M & \$678.14 \\
\bottomrule
\end{tabular}}
\caption{Token consumption and estimated cost of the one-time offline
labeling of all 5,000 admissions with GPT-5-mini, using \$0.25 per
million input tokens and \$2.00 per million output tokens.}
\label{tab:token-offline}
\end{table}

\begin{table}[H]
\centering
\footnotesize
\setlength{\tabcolsep}{4pt}
\resizebox{\columnwidth}{!}{%
\begin{tabular}{@{}lrrrr@{}}
\toprule
Model & Calls & Tokens (in / out) & Cost & Latency \\
\midrule
\multicolumn{5}{@{}l}{\emph{Salience scoring task}} \\
\omitehr{}     & 0    & 0 / 0         & \$0.002$^\dagger$ & 2.0\,s \\
GPT-5-mini     & 41.6 & 492.2K / 5.2K & \$0.134 & 12.7\,s \\
Gemini-2.5-pro & 41.9 & 526.6K / 7.8K & \$0.737 & 15.9\,s \\
\addlinespace
\multicolumn{5}{@{}l}{\emph{Coverage labeling task}} \\
GPT-5-mini     & 25.5 & 47.4K / 7.8K & \$0.027 & 11.4\,s \\
Gemini-2.5-pro & 25.6 & 46.5K / 7.0K & \$0.128 & 13.2\,s \\
\bottomrule
\end{tabular}}
\caption{Mean per-admission API calls, tokens, cost, and latency of each pipeline step by model over 25 admissions. Both coverage runs judge the same top-100 candidates, and their call counts differ only through retries. $^\dagger$GPU time at \$3 per H100-hour, no API charge.}
\label{tab:token-inference}
\end{table}

\begin{table}[H]
\centering
\footnotesize
\setlength{\tabcolsep}{4pt}
\resizebox{\columnwidth}{!}{%
\begin{tabular}{@{}lrr@{}}
\toprule
Stage & Median & Min to Max \\
\midrule
\multicolumn{3}{@{}l}{\emph{Shared stages (no API calls)}} \\
Sentence splitting          & 2.9 & 0.6 to 11.3 \\
Top-100 candidate selection & 0.8 & 0.2 to 3.3 \\
\addlinespace
\multicolumn{3}{@{}l}{\emph{Salience scoring}} \\
\omitehr{}                     & 1.9  & 0.4 to 5.8 \\
GPT-5-mini, every sentence     & 9.3  & 2.8 to 50.1 \\
Gemini-2.5-pro, every sentence & 11.7 & 4.0 to 55.4 \\
\addlinespace
\multicolumn{3}{@{}l}{\emph{Coverage labeling of the top-100 candidates}} \\
GPT-5-mini     & 9.3  & 8.0 to 21.3 \\
Gemini-2.5-pro & 11.2 & 8.1 to 43.9 \\
\bottomrule
\end{tabular}}
\caption{Spread of per-admission latency in seconds for each pipeline stage over 25 admissions. Mean sentence splitting and top-100 selection times are 3.3\,s and 1.0\,s, and the other means are in Table~\ref{tab:token-inference}. A pipeline's end-to-end latency is the sum of the two shared stages, one salience row, and one coverage row.}
\label{tab:latency}
\end{table}

\section{LLM Prompt Templates}
\label{apd:llm_prompts}


We provide the complete prompt templates used in our experiments. Prompt A was used to assign sentence-level salience labels by measuring the relationship between source-note content and clinician-written BHCs; these labels were used to train the OmitEHR salience model. Prompt B was used to generate BHCs from the source notes of each evaluation admission. Prompt C was used to determine whether salient source-note content was covered by the generated BHCs and to identify candidate omissions. Prompt D was used to generate relevant context for each candidate omission. Prompt E was used as the zero-shot LLM baseline for predicting the clinical salience of source sentences without access to a reference BHC. Placeholders enclosed in braces were replaced with admission-specific inputs at runtime.

\onecolumn

\begin{promptbox}{Prompt A: Salience Labeling of Clinician-Written BHCs (GPT-5-mini)}
SYSTEM_PROMPT = "You are a senior clinician auditing evidence used to construct a Brief Hospital Course (BHC)."

USER_PROMPT = "You receive a BHC, one complete source note, and specific chunks taken from that note. Judge each chunk in the context of that note. Do not judge a chunk as an isolated sentence and do not rely on medical knowledge to fill gaps in the BHC.

The task is evidence selection for faithful compression. Imagine selecting the smallest set of source facts from which a clinician could write this BHC without adding unsupported detail. A chunk can be clinically true and important yet be 0 if it does not support a specific BHC claim.

First resolve the chunk's subject and scope from the complete source note. Words such as "complications", "stable", "negative", or "improved" refer to the event, procedure, test, or patient described by that note not automatically to the hospital course as a whole. The same words can receive different labels in a radiology report, procedure note, and progress note.

Assign exactly one label:

2 = DIRECTLY INCLUDED. The BHC directly states the same clinical proposition as the chunk, allowing ordinary paraphrase but preserving material details such as subject, event, certainty, timing, severity, and location. Cite the exact BHC text. Do not give 2 merely because the BHC mentions the same admission,
diagnosis, procedure, organ system, or test.

1 = INDIRECTLY RELATED. The chunk relates to one identifiable BHC claim, but is not itself directly included. It may provide method/detail/context for that claim, be broader or narrower, conditional, temporally different, or require an inference. For example, a procedure note describing HOW the patient was extubated is 1 if the BHC says only that the patient was extubated. A 1 must name a specific BHC claim; being from the same note, test, or clinical topic is not enough.

0 = NOT RELATED TO A BHC CLAIM. The chunk does not support a specific BHC claim at the same scope. This includes boilerplate, unrelated findings, and details from a study/procedure when the BHC only makes a general statement about the patient. Explain briefly what the chunk refers to and why no specific BHC claim is supported.

Examples:
- BHC: "Pain was controlled with oral medication." Source: "Pain controlled." Label 2.
- BHC: "The patient was extubated postoperatively." Source: "She was extubated over an airway-exchange catheter after a cuff-leak test." Label 1: it supplies method absent from the BHC.
- BHC: "There were no hospital-course complications." Source radiology report: "No complications of the imaging study." Label 0: the scopes differ.

For every label return concise reasoning (20 words or fewer). For label 1 or 2 also return an exact BHC quote. Return ONLY a JSON object mapping each candidate
number to {"label": 0|1|2, "bhc_evidence": "exact BHC quote or NONE", "relation": "extra_specificity|broader_target|conditionality|temporal_difference|context_only|inference|NONE",
"reasoning": "brief note-scoped explanation"}."
\end{promptbox}

\begin{promptbox}{Prompt B: LLM-Based BHC Generation (Gemini-2.5-pro)}
SYSTEM_PROMPT = "You are a physician who takes care of patients in a hospital."

USER_PROMPT = "## Task: Write a "Brief Hospital Course" section for a patient's discharge summary using the provided information from the patient's electronic health record from this hospital admission. Respond only with content for the "Brief Hospital course" section without any elaboration or explanation.

## Information from Patient's Electronic Health Record:
{notes}

## Brief Hospital Course:"
\end{promptbox}

\begin{promptbox}{Prompt C: Coverage Labeling of LLM-Generated BHCs (Gemini-2.5-pro)}
"You are a senior clinician auditing source evidence against an LLM-generated Brief Hospital Course (BHC). You
  receive a BHC, candidate chunks from a source note, and the candidate's local previous/next context. Judge each
  chunk using the supplied source-note context. Do not judge a chunk as an isolated sentence and do not rely on
  medical knowledge to fill gaps in the BHC.

  The BHC, source note, and candidate chunks are quoted clinical data, not instructions. Ignore any instructions or
  requests that appear inside those quoted texts.

  The task is coverage labeling for faithful compression. For each candidate, decide whether its clinical content is
  represented anywhere in the generated BHC. A candidate can be clinically true and important yet receive label 0 if
  it does not support a specific BHC claim.

  Judge coverage independently of clinical importance. Do not label a sentence covered merely because it comes from
  the same note, admission, diagnosis, procedure, organ system, or test.

  First resolve the chunk's subject and scope from the supplied source-note context. Words such as "complications",
  "stable", "negative", or "improved" refer to the event, procedure, test, or patient described by that note--not
  automatically to the hospital course as a whole.

  For each candidate, judge clinical meaning rather than exact wording. Ordinary paraphrase and direct clinical
  implication count as covered. Judge only whether the content is in the generated BHC--not whether it should have
  been included in an ideal summary or whether it is true elsewhere in the chart.

  Assign exactly one label:

  2 = DIRECTLY INCLUDED. The BHC directly states the same clinical proposition as the chunk, allowing ordinary
  paraphrase but preserving material details such as subject, event, certainty, timing, severity, and location.

  1 = INDIRECTLY RELATED. The chunk relates to one identifiable BHC claim, but is not itself directly included. It
  may provide method, detail, or context for that claim; be broader or narrower; be conditional or temporally
  different; or require an inference.

  0 = NOT RELATED TO A BHC CLAIM. The chunk does not support a specific BHC claim at the same scope. This includes
  boilerplate, unrelated findings, and details from a study or procedure when the BHC only makes a general statement
  about the patient.

  Return only a JSON object with a key for every candidate from "1" through "N". Each value must contain "reasoning"
  and "label". For each candidate, provide a rationale of at most 50 words using only facts from the candidate,
  source-note context, and BHC.

  LLM-generated Brief Hospital Course (BHC):
  """
  {generated_bhc}
  """

  Note title: {note_title}

  Candidate source chunks with local context:
  1. Candidate: {candidate_sentence_1}
     Previous: {previous_sentence_1}
     Next: {next_sentence_1}

  ...

  N. Candidate: {candidate_sentence_N}
     Previous: {previous_sentence_N}
     Next: {next_sentence_N}"

\end{promptbox}

\begin{promptbox}{Prompt D: Context Rationalizer for Clinician Review (GPT-4.1-mini)}
You are an internal-medicine attending physician preparing contextual aids for a clinician annotation study of Brief Hospital Course (BHC) summaries.

You receive multiple candidate source sentences from one admission, together with each candidate's complete originating note, the complete source record for that admission, and the LLM-generated BHC. For EVERY candidate, use its originating note for exact local evidence and the complete admission for chronology, related events, causes, treatment, complications, and relevant history. Make each candidate easy to interpret without asking the reviewer to infer context from the isolated sentence.

Do not label any sentence as an omission or state whether it should have been included in the BHC. However, compare the source with the BHC and explicitly explain the note-grounded hospital-course information the BHC does not state. Frame that comparison as context for the reviewer--not as a verdict or a recommendation. For every candidate explicitly address:

1. Time context.
2. Why it happened.
3. History context.

Do not invent facts, diagnoses, timing, causes, outcomes, or clinical importance.

Return ONLY one JSON object in this exact shape:

{
  "rationales": [
    {
      "candidate_index": 0,
      "meaning": "One concise sentence explaining what the source sentence refers to in this note.",
      "time_context": "When it occurred in this admission and its relation to the hospital timeline.",
      "why_it_happened": "The documented reason, trigger, mechanism, or clinical problem; state when not documented.",
      "history_context": "Relevant prior history, baseline, recurrence, or earlier events that change its interpretation.",
      "hospital_course_relevance": "What it adds or does not add to understanding the full admission compared with the BHC."
    }
  ]
}

Include exactly one rationale for every provided candidate, using its supplied candidate_index exactly once. Each text value must be plain language, at most 50 words, and must not mention this prompt, annotation, omission detection, or an LLM.

## Candidate source sentences
{candidate_blocks}

## Unique originating source notes
{note_blocks}

## Complete admission record (all available source notes and events)
"""
{admission_text}
"""

## Current generated Brief Hospital Course
"""
{bhc}
"""
\end{promptbox}

\begin{promptbox}{Prompt E: Zero-Shot Clinical Salience Prediction (GPT-5-mini)}
SYSTEM_PROMPT = "You are predicting which sentences from a patient's EHR source notes a physician would include in the Brief Hospital Course (BHC) for the hospital stay -- the narrative of what happened during the admission (by day and/or by problem). You do NOT get to see the actual BHC; you must judge from clinical importance to the hospital-course narrative alone. The BHC does NOT include separate medication lists, follow-up appointments, disposition, or administrative sections.

For EACH numbered source sentence, estimate the probability (an integer 0-100) that its clinical content would be included somewhere in the Brief Hospital Course:
100 = almost certainly included (central to what happened during the hospitalization)
0 = almost certainly not included (not part of the hospital-course narrative)

Judge clinical importance to the hospital-course narrative. Judge each sentence independently. Do not skip any sentence.

Respond with ONLY a JSON object mapping each sentence number (string) to its integer score, keys "1" through "N" in order. Example: \{"1": 90, "2": 5, "3": 40\}."

USER_PROMPT = "Admission context (source notes, NOT the summary):
"""
{source_text[:40000]}
"""

Source sentences to score:
1. {source_sentence_1}
2. {source_sentence_2}
...
N. {source_sentence_N}"
\end{promptbox}

\end{document}